\documentclass[journal]{IEEEtran}
\usepackage{graphicx}
\usepackage{color}
\usepackage{multirow}%
\usepackage{amsmath,amssymb,amsfonts}%
\usepackage{mathrsfs}%
\usepackage{xcolor}%
\usepackage{textcomp}%
\usepackage{manyfoot}%
\usepackage{booktabs}%
\usepackage{algorithm}%
\usepackage{algorithmicx}%
\usepackage{algpseudocode}%
\usepackage{listings}%
\usepackage{newfloat}
\usepackage{bm}
\usepackage{subcaption}
\usepackage{colortbl}
\definecolor{maroon}{cmyk}{0,0.87,0.68,0.32}
\usepackage{makecell}
\usepackage{hyperref}

\begin{document}


\title{Improving Transferable Targeted Adversarial Attack via Normalized Logit Calibration and Truncated Feature Mixing}

\author{
        Juanjuan Weng,
        Zhiming Luo,
        Shaozi Li
\IEEEcompsocitemizethanks{
\IEEEcompsocthanksitem  J. Weng, Z. Luo (Corresponding author), and S. Li are with the Department of Artificial Intelligence, Xiamen University,
Xiamen 361005, China.
}
}

\maketitle
\begin{abstract}
This paper aims to enhance the transferability of adversarial samples in targeted attacks, where attack success rates remain comparatively low. To achieve this objective, we propose two distinct techniques for improving the targeted transferability from the loss and feature aspects. First, in previous approaches, logit calibrations used in targeted attacks primarily focus on the logit margin between the targeted class and the untargeted classes among samples, neglecting the standard deviation of the logit. In contrast, we introduce a new normalized logit calibration method that jointly considers the logit margin and the standard deviation of logits. This approach effectively calibrates the logits, enhancing the targeted transferability. Second, previous studies have demonstrated that mixing the features of clean samples during optimization can significantly increase transferability. Building upon this, we further investigate a truncated feature mixing method to reduce the impact of the source training model, resulting in additional improvements. The truncated feature is determined by removing the Rank-1 feature associated with the largest singular value decomposed from the high-level convolutional layers of the clean sample. Extensive experiments conducted on the ImageNet-Compatible and CIFAR-10 datasets demonstrate the individual and mutual benefits of our proposed two components, which outperform the state-of-the-art methods by a large margin in black-box targeted attacks. 
\end{abstract}


\section{Introduction}
\label{sec:intro}

Deep neural networks (DNNs) have experienced significant advancements in various fields over the past decade, such as image classification~\cite{simonyan2014very,he2016deep}, object detection~\cite{ren2015faster,liu2016ssd}, image segmentation~\cite{long2015fully,ronneberger2015u}. However, it has been discovered that DNNs are susceptible to adversarial attacks~\cite{goodfellow2015explaining,chen2019transferability}. By adding quasi-imperceptible adversarial perturbations into the images, attackers can cause DNNs to produce incorrect predictions. This poses a severe security risk for practical applications. Notably, the adversarial samples have high transferability, which can be directly reused to attack other unknown black-box networks without knowledge of their underlying structures. Consequently, many methods have recently been proposed for improving the transferability of adversarial samples from different aspects, such as advanced gradient optimizations~\cite{dong2018boosting,lin2020nesterov} and input data augmentations~\cite{xie2019improving,byun2022improving}. Despite advancements in black-box untargeted attacks, targeted attacks continue to pose a challenge, which involves the need to deceive black-box models into categorizing inputs within a predefined target category.

In untargeted attacks, the Cross-Entropy (CE) loss function is commonly utilized to craft adversarial samples. Nevertheless, recent studies \cite{li2020towards,zhao2021success,weng2023logit} have revealed that the CE loss is inadequate in generating transferable targeted adversarial samples due to the problem of vanishing gradient during optimization. To tackle this issue, \cite{li2020towards} proposed using the Poincar{\'e} distance to adaptively amplify the gradient magnitude. ~\cite{zhao2021success} introduced the Logit loss, which is equal to the negative value of the target class's logits, ensuring a constant gradient for the targeted class. Expanding on this, \cite{weng2023logit} further found that increasing the logit margin between targeted and untargeted classes can improve the targeted transferability of adversarial samples. Subsequently, three different logit calibration methods (\textit{i.e.}, Temperature-based, Margin-based, and Angle-based) were introduced to achieve this objective. However, we also notice that the optimal calibrations vary across different source CNNs, which neglects the practical utility of logit calibration when using a new surrogate model for training.

To address the aforementioned issue, we begin by visualizing the density distribution of the logit margin and the standard deviation with logits across the ImageNet-Compatible dataset. Based on visualization, we argue that the main reason primarily arises from the variation of logit margin among samples within the same model and the variation of standard deviations between different samples and models. However, previous logit calibrations primarily focus on optimizing the logit of the targeted class and neglect the overall influence of the logit distribution. In this study, we propose a novel normalized logit calibration method that takes into account both the logit margin and the standard deviation of logit distribution.

Meanwhile, \cite{byun2023introducing} proposed clean feature mixing (CFM) to greatly enhance the targeted transferability
by randomly
mixing the stored clean features of images into the adversarial features within a batch.  In this study, we further found that only mixing the truncated clean features can have better performance than using the original clean features. Building on the findings in Eigen-CAM~\cite{muhammad2020eigen} and Rank-Feat~\cite{song2022rankfeat} that the Rank-1 feature, associated with the largest singular value, contributes most to the prediction of a source CNN model, we hypothesize that mixing features that are less correlated to the source training model could further improve the transferability. Therefore, we introduce the truncated feature mixing (TFM) by removing the Rank-1 feature from the original clean features, thereby reducing the impact from the source model to some extent.  

Finally, we conducted extensive experiments to evaluate the performance of our proposed method on the ImageNet-Compatible and CIFAR-10 datasets using various models and training baselines. Our proposed normalized logit calibration can surpass the other logit calibrations, and the proposed truncated feature mixing outperforms the CFM. The mutual benefits of these two components further demonstrate that our method surpasses state-of-the-art comparison methods by a large margin.

In summary, the main contributions of this study are as follows:
\begin{itemize}
    \item We introduce a normalized logit calibration method by jointly considering the logit margin and logit distribution of each sample for crafting the targeted adversarial examples, which is compatible with various different CNNs models.

    \item We present a truncated feature mixing strategy by removing the Rank-1 feature associated with the largest singular value, which can provide extra benefits for increasing the targeted transferability.

    \item Extensive experiments on two benchmark datasets have validated the effectiveness of our proposed method, showcasing its ability to outperform other comparative methods. Furthermore, it can be seamlessly integrated into various attacking baselines to enhance overall performance.
    
\end{itemize}

The rest of the paper is organized as follows. We review the previous untargeted and targeted black-box attacks in Section~\ref{sec:related works}. We describe the details of our proposed method in Section~\ref{sec:method}. The experimental results are reported and discussed in Section~\ref{sec:experiments}. Finally, we conclude our paper in Section~\ref{sec:conclude}.

\section{Related Works}
\label{sec:related works}

In this section, we will provide an overview of previous studies, focusing on two key aspects: untargeted black-box attacks and targeted black-box attacks.

\subsection{Untargeted Black-Box Attacks}
In the past few years, various adversarial attack methods~\cite{dong2018boosting,dong2019evading,weng2023exploring} have been proposed for crafting more destructive adversarial samples from different aspects. The Fast Gradient Sign Method (FGSM)~\cite{goodfellow2015explaining} is one of the most fundamental attack methods, which employs a one-step update to optimize perturbations as follows:
\begin{equation}
 \hat{x}=x+\epsilon\cdot\text{sign}(\nabla_{{x}}  \mathcal{L}(f(x),y) ,
\end{equation}
where $\hat{x}$ denotes the adversarial image, $\epsilon$ ensures the adversarial sample is constrained within an upper-bound perturbation through the $l_p$-norm, and ${\mathcal{L}}$ is usually the cross-entropy (CE) loss. 

The Iterative-FGSM (I-FGSM)~\cite{kurakin2018adversarial} further extends the FGSM by updating the image iteratively with a smaller step size in the gradient direction as follows:
\begin{equation}
   {\hat{x}}_{0} = x, \quad {\hat{x}}_{i+1} = {\hat{x}}_{i} + \alpha \cdot \text{sign}(\nabla_{\hat{x}_i}  {\mathcal{L}}(f(\hat{x}_{i}),y)),
   \label{eq:ifgsm}
\end{equation}
where $\alpha=\epsilon/T$ ensures that $x'$ is within the $l_p$-norm when optimized by a maximum of $T$ iterations. Moreover, adversarial samples exhibit black-box transferability in which adversarial examples obtained from one source model can be directly reused to mislead other unknown models. Various techniques have been proposed to improve transferability mainly from the following two aspects.

\textbf{Enhancing gradient stability methods} aim to increase transferability by preventing adversarial examples from getting trapped in local optima. For example, the Momentum Iterative FGSM (MI-FGSM) \cite{dong2018boosting} integrates a momentum term into the I-FGSM, promoting stable optimization. The Variance Tuning (VT) method \cite{wang2021enhancing} stabilizes the update direction by adopting the gradient
information in the neighborhood of the previous data point to tune the gradient of the current point. Similarly, the Scale-Invariant (SI) attack method \cite{lin2020nesterov} progressively scales down pixel values of the input image over multiple steps and computes gradients from this set of images.

\textbf{Input transformation-based methods} employ data augmentation strategies at the input level to enhance transferability. The Diverse-Inputs (DI) method \cite{xie2019improving} implements randomized expansion and padding of the input during each iteration. Translation-Invariant (TI) \cite{dong2019evading} technique seeks a level of translation invariance, enhancing transferability by approximating a weighted average of gradients derived from an ensemble of translated images within a specific range. The Resized Diverse-Inputs (RDI) method \cite{zou2020improving} extends the DI approach by reducing the expanded image back to its original size after the DI transformation. The Admix method \cite{wang2021admix} enhances transferability by mixing images from various categories within the image domain, which computes gradients on the input image mixed with a fraction of each additional image, while maintaining the original input's label. The S$^2$I-FGSM \cite{long2022frequency} proposes a frequency domain data augmentation for training by applying a spectrum transformation to the input.

It is worth noting that many targeted attack methods generally combine the aforementioned gradient stabilization techniques and input transformation-based approaches to form a strong baseline with better transferability. In this study, we evaluate the performance of our proposed method under various baselines.

\subsection{Targeted Black-Box Attacks}
Different from non-targeted attacks~\cite{xie2019improving,huang2019enhancing}, targeted attacks focus on misleading deep models into a predefined target class, which is a more challenging task. A majority of studies~\cite{weng2023logit,zhao2021success} have primarily focused on black-box targeted attacks. These studies utilize the transferability of adversarial examples to attack unknown deep models, leading them to produce incorrect predictions aligned with the target labels. 

In targeted attacks, the previously discussed techniques in non-targeted attacks also can be used to improve transferability. Additionally, the Object-based Diverse Input (ODI) method~\cite{byun2022improving} is a recent input transformation-based approach by projecting an input image onto a randomly selected 3D object's surface and then displaying this painted object in various rendering environments. Moreover, there are two additional aspects of techniques aimed at improving targeted transferability.

\textbf{Loss Function-based methods} focus on innovating new loss functions to enhance targeted transferability, aiming at alleviating the gradient saturation issue in the conventional cross-entropy loss. The Po+Trip method~\cite{li2020towards} introduces the Poincaré distance as a similarity metric, dynamically amplifying gradient magnitudes through Triplet loss to guide adversarial examples towards the target label. \cite{zhao2021success} introduce the logit loss, which is equal to the negative value of the target class's logits, guaranteeing a consistent gradient for the target class. \cite{weng2023logit} build on the Logit~\cite{zhao2021success}and identify that increasing the logit margin between targeted and untargeted classes enhances targeted transferability. Consequently, it introduces three distinct logit calibration methods (Temperature-based, Margin-based, and Angle-based) to achieve this objective. 

\textbf{Feature-based methods} leverage the intermediate features of adversarial images to enhance targeted attack performance. The FDA method~\cite{inkawhich2020transferable,inkawhich2020perturbing} improves targeted transferability by utilizing intermediate feature distributions extracted from CNNs. This is accomplished through training class-specific auxiliary classifiers that model layer-wise feature distributions. ~\cite{gao2021feature} introduced Pair-wise Alignment Attack and Global-wise Alignment Attack to explore feature map similarities using high-order statistics with translation invariance. The CFM method~\cite{byun2023introducing} aims to elevate the targeted transferability of adversarial examples through efficient simulation of competitor noises, which is achieved by randomly mixing the stored clean features of images within a batch.

\section{Method}
\label{sec:method}
\subsection{Problem Definition}
The primary objective of targeted attacks is to learn an adversarial perturbation $\delta$ for an input image $x$ with the aim of misleading deep models into predicting a specific target class $t$. To achieve this objective, a white-box model $f$ is employed to craft the targeted adversarial sample $\hat{x} = x + \delta$.
Subsequently, we utilize the transferability of $\hat{x}$ to deceive unknown black-box models, inducing them to generate incorrect predictions that align with the target labels $t$. Besides, the perturbation $\delta$ is usually bounded within an upper-bound $\epsilon$ by the commonly used $l_\infty$-norm, denoted as $||\delta||_\infty  \leq \epsilon$. 

In a nutshell, the learning objective loss function for the targeted adversarial attack can be formally denoted as:
\begin{equation}
    \arg\min\limits_{\hat{x}} \mathcal{L}\left(f(\hat{x}), t\right),\quad || \hat{x} - x ||_\infty \leq \epsilon,
\end{equation}
where $\mathcal{L}$ is the training adversary loss function for the targeted attack. During the training procedure, the I-FGSM optimization method (Eq.~\ref{eq:ifgsm}) is usually used to optimize the targeted adversarial examples.

\begin{figure}[t]
    \centering
    \begin{subfigure}{0.48\linewidth}
    \includegraphics[width=\linewidth]{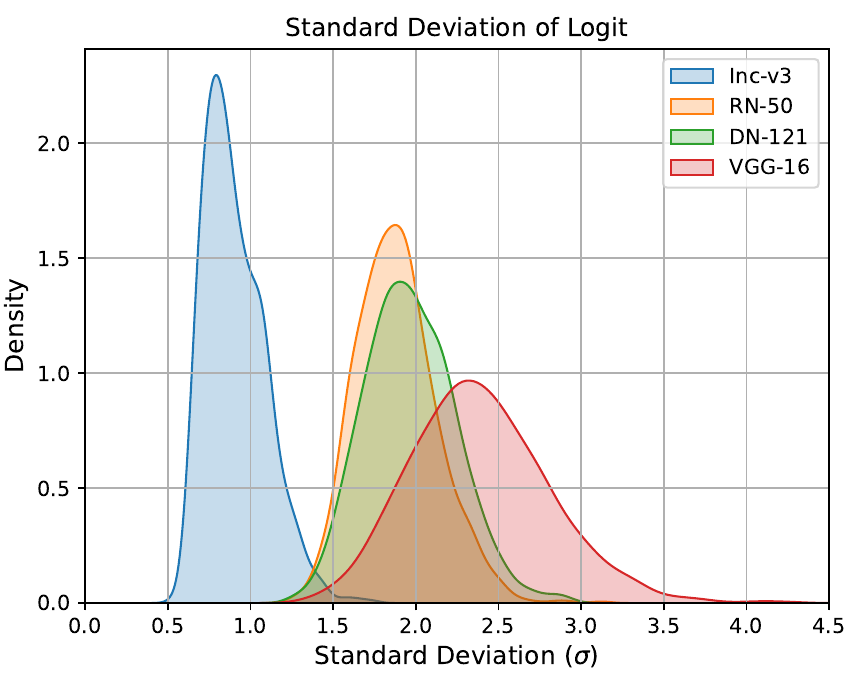}
        \label{fig:std_new}
    \end{subfigure}
    \begin{subfigure}{0.48\linewidth}
   \includegraphics[width=\linewidth]{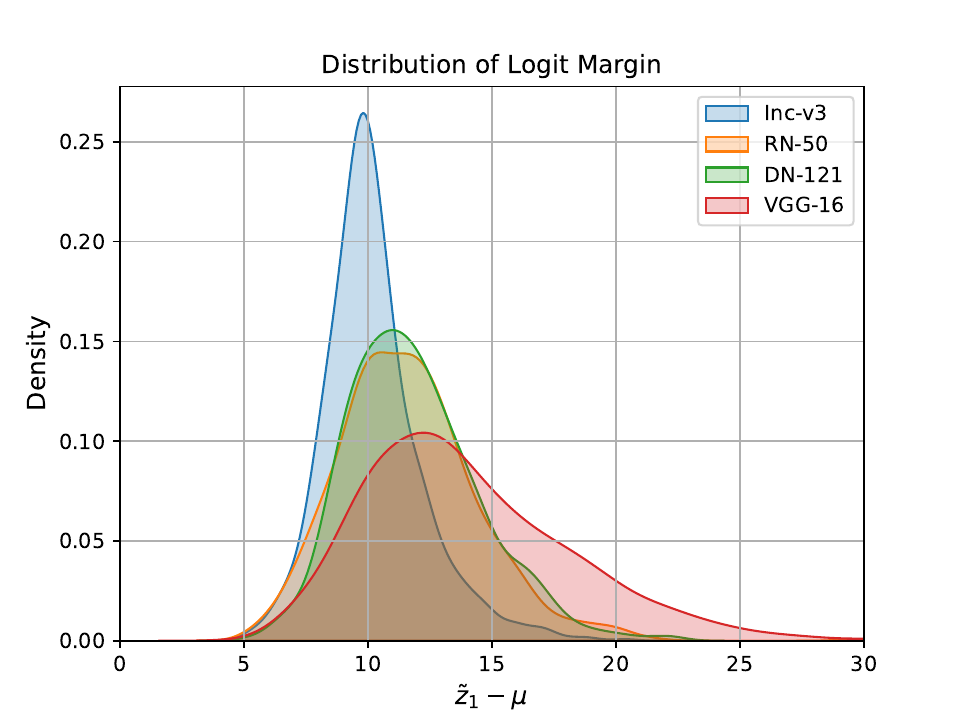}
        \label{fig:mean_new}
    \end{subfigure}
    \caption{The density distributions of the standard deviation of logits and the logit margins obtained from the 1000 images across the ImageNet-Compatible dataset.}
    \label{fig:logits_density}
\end{figure}

\subsection{Normalized Logit Calibration \& Loss}
\subsubsection{Revisiting Logit Loss and Logit Calibration}
In this section, we start by revisiting the Logit Loss \cite{zhao2021success} and Logit Calibrations \cite{weng2023logit} that aim to craft transferable targeted adversarial samples. Firstly, we denote the logit value before the final softmax function as $z_i$ for each class $i$ and the logit distribution of $x$ as $\bm{z}=\{z_1, z_2,..., z_C\}$. The $C$ is the total number of categories. After sorting $\bm{z}$ in the descend order, we can have $\Tilde{\bm{z}} = \{\Tilde{z}_1, \Tilde{z}_2,..., \Tilde{z}_C\}$.

The Logit Loss proposed in~\cite{zhao2021success} for the targeted attack is defined as the negative value of the target class's logit:
\begin{equation}
    \mathcal{L}_{logit} = -z_t,
\end{equation}
where $z_t$ is the logit value of adversarial example $\hat{x}$ corresponding to the target class $t$.
This Logit loss can effectively address the vanishing gradient issue in the original cross-entropy loss since the gradient $\frac{\partial L_{logit}}{\partial z_t}$ will always be a constant that equals to -1. Experimental results verified the effectiveness of $\mathcal{L}_{logit}$ in improving targeted transferability when training for more iterations (e.g., $iters=300$).

Building upon the findings in \cite{zhao2021success}, our previous study~\cite{weng2023logit} further observed that increasing the logit margin between targeted and untargeted classes can enhance targeted transferability. Therefore, three different logit calibration methods have been introduced to achieve this goal, which are described as follows:
(1) Temperature-based: $\hat{z_i} = \frac{z_i}{T};$
(2) Margin-based: $\hat{z_i} = \frac{z_i}{\Tilde{z}_1 - \Tilde{z}_2};$
(3) Angle-based: $\hat{z_i} = \frac{z_i}{||\bm{w}_i|||\phi(\hat{x})||}$, where $\Tilde{z}_1$ and $\Tilde{z}_2$ are rank-1 and rank-2 logit value of adversarial example, $\bm{w}_i$ is weights in classification layer for the class $i$, and $\phi(\hat{x})$ is the final feature for computing logits.

Subsequently, the cross-entropy loss function is applied for training with Temperature-based and Margin-based calibrations, and the Logit loss is used for Angle-based calibration. However, based on the experimental results in~\cite{weng2023logit}, we can notice the optimal calibrations and values of $T$ vary for different surrogate models. For example, Temperature-based calibration is more suitable for ResNet-50 ($T=5$) and DenseNet-121 ($T=10$), while Margin-based calibration is preferred for VGG-16. Consequently, this neglects the practical utility of logit calibration for the targeted attack when using a new surrogate model for training which needs to select the optimal calibrations.

To further analyze the issues discussed in the Logit Calibrations \cite{weng2023logit}, we visually present the density distribution of logit margin $\Tilde{z}_1 - \mu$ and the standard deviation $\sigma$ of logit distribution $\bm{Z}$ calculated from multiple models using the 1000 samples from the ImageNet-Compatible dataset. Figure~\ref{fig:logits_density} shows the visualization of these distributions. From the figure, several observations can be made: \textbf{1)} The logit margin $\Tilde{z}_1 - \mu$ exhibits significant variation across distinct samples within the same model, indicating that the model's predictions can differ greatly for individual input instances. However, this variation tends to be relatively consistent across different source models. \textbf{2)} The standard deviations $\sigma$ of logit distribution also show disparity among different CNNs. Some CNNs exhibit higher standard deviations than others, suggesting that they have a broader range of prediction probabilities for different classes.
Interestingly, CNNs with similar distributions, such as ResNet-50 and DenseNet-121, may exhibit better transferability between each other,  as indicated in the results presented in the experiment section. This implies that models with similar logit distributions are more likely to have more shared decision boundaries and therefore enhance the transferability of targeted adversarial samples between them.
\textit{Based on these observations, we raise the assumption that by reducing the variations of different samples within the same model and increasing the similarity between different models, we can potentially improve the transferability of targeted adversarial samples. }

\subsubsection{Normalized Logit Calibration}


To reduce the variation in the logits $Z$, a straightforward solution could involve treating the logit distribution over the $C$ different categories of each sample as a Gaussian distribution. This can be achieved by normalizing the logits to the standard distribution with an extra temperature factor, resulting in $\hat{z}_i = \frac{z_i - \mu}{\sigma \times T}$. In our experiments, we observed that this Gaussian-based logit calibration method can improve the targeted attack success rates and outperform the Logit loss (Table~\ref{tab:calibration}). However, its performance is lower than the Temperature-based and Margin-based calibrations proposed in \cite{weng2023logit}. 

By analyzing the corresponding cross-entropy loss function, expressed as:
\begin{align}
        \mathcal{L}_{CE}  &= -\log \left(\frac{\exp(\frac{z_t - \mu}{\sigma \times T})}{\sum_i \exp(\frac{z_i - \mu}{\sigma \times T})}\right) \\\nonumber
                &= -\log \left(\frac{\exp(\frac{z_t}{\sigma \times T})/\exp(\frac{\mu}{\sigma \times T})}{\sum_i \exp(\frac{z_i}{\sigma \times T})/\exp(\frac{\mu}{\sigma \times T})} \right) \\ \nonumber
                & = -\log \left(\frac{\exp(\frac{z_t}{\sigma \times T})}{\sum_i \exp(\frac{z_i}{\sigma \times T})}\right),
\end{align}
we observe that $\hat{z}_i = \frac{z_i - \mu}{\sigma \times T}$ is equivalent to the calibration $\hat{z}_i = \frac{z_i}{\sigma \times T}$, where only the standard deviation $\sigma$ is considered without accounting for the logit margin. During the learning iterations, the $\sigma$ will keep increasing along with the logit value related to the target class. As a result, the $\sigma$ will over-smooth the logits, which thus leads to a negative influence on the performance. 

On the other hand, we are encouraged to enlarge the logit margin during the optimization, thus increasing the separation between the targeted class and other non-targeted classes to improve transferability. The logit margin can be referred to as the margin $z_t - \mu$ between the logit of the targeted class and the mean value of logits. Instead of normalizing the logit distribution of each sample to the Gaussian distribution without considering the logit margin, we propose jointly utilizing the logit margin and standard deviation for calibration. The corresponding logit calibration factor can be calculated as:
\begin{equation}
    C = \frac{\Tilde{z}_1 - \mu}{\sigma \times T}.
    \label{eq:ncl}
\end{equation}
The $\Tilde{z}_1$ will be $z_t$ after optimizing with a few iterations. Subsequently, we calibrate the logits for each sample by $\hat{z_i} = \frac{z_i}{C}$.

\subsubsection{Loss Function}
After calibrating the logits of each sample, we have the final normalized cross-entropy (NCE) loss function for training targeted adversarial samples as follows:
\begin{equation}
    L_{NCE} = -\log \left(\frac{\exp(\hat{z}_t)}{\sum_i \exp(\hat{z}_i)} \right).
\end{equation}


\begin{figure}[t]
    \centering
    \includegraphics[width=\linewidth]{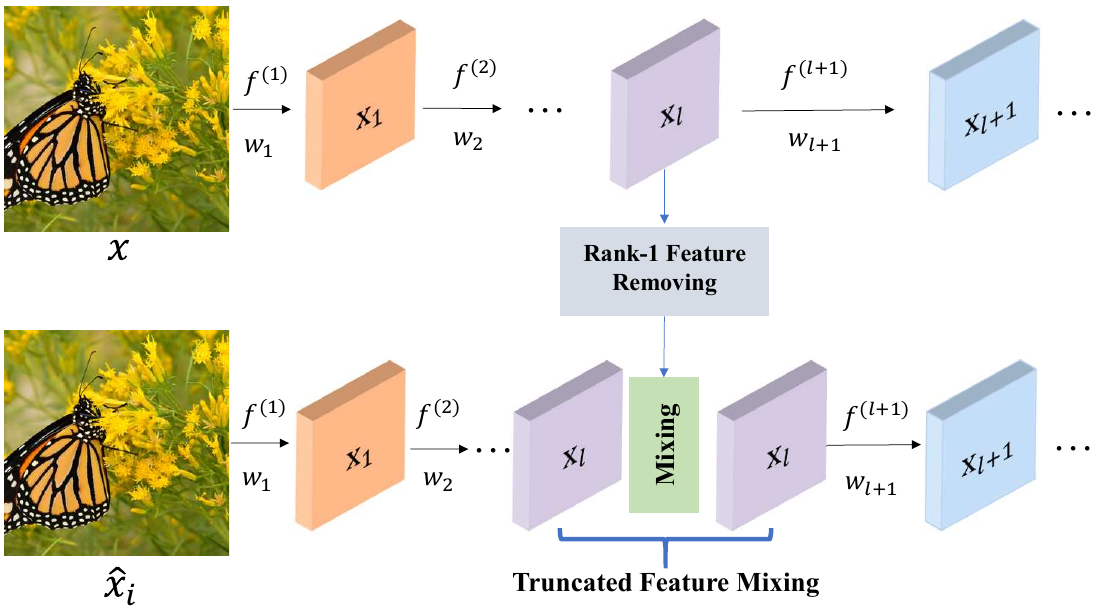}
    \caption{Overview of the Truncated Feature Mixing procedure.}
    \label{fig:tfm}
\end{figure}

\subsection{Truncated Feature Mixing}

Eigen-CAM~\cite{muhammad2020eigen} and RankFeat~\cite{song2022rankfeat} have shown that the Rank-1 feature, associated with the largest singular value of a high-level convolutional layer's feature, contributes most to the prediction for a source CNN model. This indicates a higher correlation between the Rank-1 feature and the real ground-truth class. Since the main objective of this study is to enhance targeted transferability, we assume that reducing the impact on the Rank-1 feature related to the source model will enhance transferability. Consequently, we achieve this by amplifying information from other singular values to execute the attack, thereby diminishing the influence of the Rank-1 feature. Therefore, we propose a Truncated Feature Mixing to achieve this goal.
The overall computation procedure of this step is illustrated in Fig.~\ref{fig:tfm}, which mainly consists of two steps, \textit{i.e.}, Rank-1 Feature Removing, and Feature Mixing. In the following sections, we will describe the computation details. 

\subsubsection{Rank-1 Feature Removing} 
Given a clean sample $x$, we can obtain the feature map $\bm{X}\in R^{C \times H \times W}$ from a high-level convolutional layer $l$. Here `high-level' denotes the feature map with rich semantics, which usually are those convolutional layers after two pooling layers. Then, we perform a reshape operation to convert the $\bm{X}$ into the size $\bm{X} \in R^{C \times HW}$. Next, we leverage the Singular Value Decomposition (SVD) to decompose the feature $\bm{X}$:
\begin{equation}
    \bm{X} = \bm{U}\bm{S}\bm{V}^T,
\end{equation}
where $\bm{S}$ is the diagonal singular value matrix, $\bm{U}$ is the left orthogonal singular vector matrix, and $\bm{V}$ is the right orthogonal singular vector matrix.

After the feature decomposition, we calculate the Rank-1 feature associated with the largest singular value $s_1$ and remove it from the original feature $\bm{X}$. This removing step can be denoted as:
\begin{equation}
    \bm{X}^{'} = \bm{X} - s_{1} \bm{u}_{1} \bm{v}_{1}^{T},
\end{equation}
where $\bm{u}_1$ and $\bm{v}_1$ are the corresponding left and right singular vectors, respectively. The resulting $\bm{X}^{'}$ is then used for the feature mixing to enhance the targeted transferability.

\subsubsection{Feature Mixing} 
Following the Clean Feature Mixup used in~\cite{byun2023introducing}, we perform the feature mixing between the truncated feature $\bm{X}^{'}$ of the original clean sample $x$ and the feature $\bm{\hat{X}}$ of the corresponding adversarial sample $\hat{x}$ at the same layer. During the optimization iteration, the feature mixing is conducted with a
probability $p$, denoted as:
\begin{equation}
    \bm{\hat{X}} = \bm{\alpha} \odot \bm{\hat{X}} + (1-\bm{\alpha}) \odot \bm{X}^{'}.
\end{equation}
Here, $\bm{\alpha} \in R^{C\times 1\times 1}$ is the channel-wise mixing ratio, which is randomly sampled from a uniform distribution $\bm{\alpha} \sim U(0,\alpha_{max} )$, and $0 \leq \alpha_{max} \leq 1$.

Besides, the \textbf{Random Feature Shuffle} is also adopted during the feature mixing, which randomly shuffles the $\bm{X}^{'}$ within a training batch. Then, the $\hat{\bm{X}}$ can mix the feature $\bm{X}^{'}$ either from itself or other images. Notice that, instead of performing the Mixup at both convolutional layers and final linear layers in CFM~\cite{byun2023introducing}, we only conduct the feature mixing at the convolutional layers.


\begin{table*}[ht!]
\caption{Targeted attack success rates(\%) against ten target models on the ImageNet-Compatible dataset.}
\centering
\label{tab:non-robust}
\begin{tabular}{lccccccccccc}
\hline
\textbf{Source: RN-50} & \multicolumn{9}{c}{Target model} & \multicolumn{1}{l}{} \\
 \cline{2-11}
Attack & VGG-16 & RN-18 & RN-50 & DN-121 & Xcep & MB-v2 & EF-B0 & IR-v2 & Inc-v3 & Inc-v4 & Avg. \\ \hline
DI          &62.5 &56.6 &98.9 &72.3 & 5.7 &28.2 &29.3 & 4.5 & 9.2 & 9.9 &37.7 \\
RDI         &65.4 &71.8 &98.0 &81.3 &13.1 &46.6 &46.6 &16.8 &30.7 &23.9 &49.4\\
SI-RDI      &70.5 &79.8 &98.8 &88.9 &29.5 &56.2 &66.2 &37.9 &56.4 &43.6 &62.8\\
VT-RDI      &68.8 &78.7 &98.2 &82.5 &27.9 &54.5 &56.1 &32.8 &45.8 &37.9 &58.3\\
Admix-RDI   &74.2 &80.7 &98.7 &86.8 &20.9 &59.4 &56.1 &26.7 &42.7 &34.1 &58.0\\
ODI         &78.3 &77.1 &97.6 &87.0 &43.8 &67.3 &70.0 &49.5 &65.9 &55.4 &69.2\\
CFM-RDI     &84.7 &88.4 &98.4 &90.3 &51.1 &81.5 &78.8 &48.0 &65.5 &59.3 &74.6\\
\rowcolor[gray]{.93}
TFM-RDI    &87.1	&88.5	&98.7	&90.6	&50.7	&83.1	&79.7	&47.7	&66.4	&61.9	&75.4\\
\rowcolor[gray]{.93}
TFM-RDI+NCE       & 91.0 & 93.8 & 100.0 & 95.9 & 56.1 & 89.9 & 85.8 & 52.9 & 72.5 & 68.3 & 80.6
\\

\hline
\textbf{Source: adv-RN-50} & \multicolumn{9}{c}{Target model} & \multicolumn{1}{l}{} \\
 \cline{2-11}
Attack & VGG-16 & RN-18 & RN-50 & DN-121 & Xcep & MB-v2 & EF-B0 & IR-v2 & Inc-v3 & Inc-v4 & Avg. \\ \hline
DI          &65.3	&81.5	&91.5	&87.0	&32.6	&62.5	&68.8	&36.9	&55.3	&42.2	&62.4 \\
RDI         &59.7	&83.5	&90.7	&85.9	&39.7	&67.0	&68.8	&44.2	&62.4	&45.1   &64.7\\
SI-RDI      &53.9	&79.4	&87.1	&83.8	&46.6	&66.5	&69.5	&52.0	&69.1	&52.2	&66.0\\
VT-RDI      &54.0	&76.8	&84.7	&81.2	&38.5	&60.3	&58.7	&42.7	&56.1	&44.9	&59.8\\
Admix-RDI   &62.7	&83.0	&90.3	&86.6	&46.9	&71.8	&72.4	&48.8	&66.3	&53.0	&68.2\\
ODI         &62.0	&77.6	&84.3	&85.0	&56.3	&66.9	&73.0	&61.1	&71.9	&60.0	&69.8\\
CFM-RDI     &76.7	&86.3	&90.9	&87.6	&67.1	&82.4	&83.4	&64.7	&77.1	&67.4	&78.4\\
\rowcolor[gray]{.93}
TFM-RDI &79.8	&87.9	&92.4	&89.8	&67.5	&84.6	&85.3	&67.4	&79.7	&69.9	&80.4\\
\rowcolor[gray]{.93}
TFM-RDI+NCE       &85.7 & 93.2 & 95.4 & 94.4 & 73.3 & 89.3 & 89.7 & 70.9 & 83.5 & 75.3 & 85.1\\

\hline
\textbf{Source: DN-121} & \multicolumn{9}{c}{Target model} & \multicolumn{1}{l}{} \\
 \cline{2-11}
Attack & VGG-16 & RN-18 & RN-50 & DN-121 & Xcep & MB-v2 & EF-B0 & IR-v2 & Inc-v3 & Inc-v4 & Avg. \\ \hline

DI          &37.4	&28.7	&44.4	&98.7	&5.2	&13.1	&18.7	&4.3	&7.1	&8.3	&26.6\\
RDI         &42.1	&48.8	&55.7	&98.5	&10.1	&21.0	&29.0	&12.8	&20.8	&18.8	&35.8\\
SI-RDI      &45.4	&53.0	&60.1	&98.6	&16.1	&27.8	&37.3	&22.0	&34.3	&25.8	&42.0\\
VT-RDI      &47.7	&56.7	&62.1	&98.6	&20.3	&28.7	&36.9	&25.4	&31.5	&27.2	&43.5\\
Admix-RDI   &53.2	&60.7	&67.6	&98.3	&17.8	&31.5	&39.4	&20.1	&31.1	&26.5	&44.6\\
ODI         &64.2	&64.2	&71.7	&98.0	&31.4	&45.9	&56.1	&39.8	&52.8	&45.9	&57.0\\
CFM-RDI     &76.2	&79.0	&83.9	&97.8	&41.1	&62.5	&68.6	&43.6	&56.1	&53.8	&66.3\\
\rowcolor[gray]{.93}
TFM-RDI &76.2	&80.7	&84.3	&98.1	&42.4	&64.2	&69.1	&44.6	&59.9	&53.8	&67.3\\
\rowcolor[gray]{.93}
TFM-RDI+NCE     & 85.7 & 89.6 & 92.5 & 100.0 & 48.2 & 73.4 & 78.0 & 50.6 & 67.4 & 62.5 & 74.8\\
\midrule[1.2pt]
\textbf{Source:Inc-v3} & \multicolumn{9}{c}{Target model} & \multicolumn{1}{l}{} \\
 \cline{2-11}
Attack & VGG-16 & RN-18 & RN-50 & DN-121 & Xcep & MB-v2 & EF-B0 & IR-v2 & Inc-v3 & Inc-v4 & Avg. \\ \hline

DI          &2.9	&2.4	&3.4	&5.0	&1.9	&1.8	&3.7	&3.0	&99.2	&4.2	&12.8\\
RDI         &3.5	&3.8	&4.0	&7.0	&3.1	&3.0	&5.9	&6.3	&98.7	&7.1	&14.2\\
SI-RDI      &4.0	&5.2	&5.7	&11.0	&6.3	&4.6	&8.2	&11.6	&98.8	&12.1	&16.8\\
VT-RDI      &5.9	&8.9	&9.4	&13.2	&7.4	&5.9	&9.8	&12.3	&98.7	&14.7	&18.6\\
Admix-RDI   &6.3	&6.5	&8.8	&12.8	&6.0	&6.1	&10.9	&12.2	&98.7	&13.6	&18.2\\
ODI         &14.3	&14.9	&16.7	&32.3	&20.3	&13.7	&25.3	&26.4	&95.6	&31.6	&29.1\\
CFM-RDI     &22.9	&26.8	&26.2	&39.1	&34.1	&27.1	&38.6	&36.2	&95.9	&44.8	&39.2\\
\rowcolor[gray]{.93}
TFM-RDI & 22.1 & 26.6 & 27.4 & 39.1 & 39.4 & 23.8 & 39.1 & 44.9 & 96.8 & 49.4 & 40.9 \\

\rowcolor[gray]{.93}
TFM-RDI+NCE & 21.6	& 27.4	& 28.8	& 38.5	& 36.8	& 23.1	& 40.3	& 44.8	& 96.0	& 48.3	& 40.6\\

\hline
\end{tabular}
\end{table*}

\begin{table*}[ht!]
\caption{Targeted attack success rates (\%) against a robust model and five Transformer-based models on the ImageNet-Compatible dataset.}
\centering
\label{tab:vit}
\begin{tabular}{lccccccc}
\hline
\textbf{Source: RN-50} & \multicolumn{6}{c}{Target model} & \multicolumn{1}{l}{} \\
 \cline{2-7}
Attack & \makecell{adv-RN-50} & ViT & LeViT & ConViT & Twins & PiT & Avg. \\ \hline
DI     & 10.9 & 0.1	& 3.6  & 0.3 & 1.3	& 1.5 & 3.0 \\
RDI    & 34.8 & 0.7	& 13.1 & 1.9 & 5.9	& 6.8 & 10.5  \\
SI-RDI & 59.9	& 2.9	& 29.4	& 6.3	& 15.5	& 17.9	& 22.0 \\
VT-RDI  & 64.2	& 2.9	& 28.1	& 5.2	& 15.0	& 14.0	& 21.6 \\
Admix-RDI   & 52.4	& 1.3	& 22.5	& 2.5	& 8.5	& 8.4	& 15.9 \\
ODI         & 64.7	& 5.1	& 37.0	& 10.7	& 20.1	& 29.1	& 27.8 \\
CFM-RDI  & 75.5	& 4.3	& 46.1	& 8.9	& 25.2	& 24.7	& 30.8\\
\rowcolor[gray]{.93}
TFM-RDI &75.9	&4.4	&47.0	&8.8	&25.1	&27.6	&31.5\\
\rowcolor[gray]{.93}
TFM-RDI+NCE       & 82.1 & 4.5 & 52.4 & 11.1 & 26.9 & 30.6 & 34.6\\


\hline
\textbf{Source: adv-RN-50} & \multicolumn{6}{c}{Target model} & \multicolumn{1}{l}{} \\
 \cline{2-7}
Attack & \makecell{adv-RN-50} & ViT & LeViT & ConViT & Twins & PiT & Avg. \\ \hline
DI     & 98.9 & 5.7	 & 36.9	& 10.1	& 19.2 & 20.5 & 31.9 \\
RDI    & 98.8 & 10.8 & 49.5	& 19.9	& 29.4 & 35.8 & 40.7 \\
SI-RDI & 98.7	& 19.4	& 57.6	& 35.3	& 35.2	& 52.1	& 49.7 \\
VT-RDI  & 98.5	& 10.6	& 46.3	& 20.0	& 27.1	& 34.4	& 39.5 \\
Admix-RDI   & 98.9	& 12.1	& 55.5	& 23.1	& 32.4	& 38.9	& 43.5 \\
ODI     & 97.3	& 22.2	& 57.7	& 38.8	& 40.0	& 54.9	& 51.8 \\
CFM-RDI  & 98.3	& 29.5	& 69.8	& 41.8	& 52.7	& 59.8	& 58.7 \\
\rowcolor[gray]{.93}
TFM-RDI &98.4	&30.0	&70.8	&45.4	&54.1	&63.4	&60.4\\
\rowcolor[gray]{.93}
TFM-RDI+NCE        & 99.6 & 32.6 & 76.7 & 47.5 & 59.4 & 68.1 & 64.0\\

\hline
\textbf{Source: DN-121} & \multicolumn{6}{c}{Target model} & \multicolumn{1}{l}{} \\
 \cline{2-7}
Attack & \makecell{adv-RN-50} & ViT & LeViT & ConViT & Twins & PiT & Avg. \\ \hline
DI     & 3.2  & 0.2 & 3.0  & 0.4 & 1.0	& 1.1 & 1.5    \\
RDI    & 10.1 & 0.8	& 8.5  & 1.3 & 3.7	& 4.5 & 4.8 \\
SI-RDI & 19.2	& 2.0	& 16.1	& 2.4	& 8.2	& 11.7	& 9.9 \\
VT-RDI & 26.6	& 2.2	& 19.2	& 3.5	& 8.3	& 11.7	& 11.9 \\
Admix-RDI   & 19.2	& 1.0	& 14.7	& 1.7	& 6.8	& 7.4	& 8.5 \\
ODI    & 35.6	& 3.3	& 26.9	& 7.4	& 14.7	& 21.9	& 18.3 \\
CFM-RDI  & 43.2	& 3.6	& 32.8	& 6.4	& 17.3	& 21.1	& 20.7 \\
\rowcolor[gray]{.93}
TFM-RDI &43.5	&2.7	&34.8	&6.1	&19.3	&21.4	&21.3\\
\rowcolor[gray]{.93}
TFM-RDI+NCE        & 50.7 & 3.1 & 41.8 & 7.2 & 22.3 & 23.7 & 24.8\\


\hline
\textbf{Source: Inc-v3} & \multicolumn{6}{c}{Target model} & \multicolumn{1}{l}{} \\
 \cline{2-7}
Attack & \makecell{adv-RN-50} & ViT & LeViT & ConViT & Twins & PiT & Avg. \\ \hline
DI    & 0.2	& 0.1 & 0.3	& 0.0	  & 0.0	& 0.1 & 0.1 \\
RDI   & 0.8	& 0.2 & 1.8	& 0.2 & 0.4	& 0.7 & 0.7 \\
SI-RDI & 2.0	& 0.3	& 4.1	& 0.9	& 0.7	& 3.2	& 1.9 \\
VT-RDI  & 3.2	& 0.4	& 5.2	& 0.8	& 1.6	& 1.8	& 2.2 \\
Admix-RDI & 2.0 & 0.1	& 4.1 & 0.6	& 1.4 & 1.4 & 1.6 \\
ODI  & 6.5	& 0.8	& 12.4	& 1.7	& 3.5	& 6.7	& 5.3 \\
CFM-RDI  & 8.6	& 2.1	& 21.9	& 3.2	& 6.1	& 11.6	& 8.9 \\
\rowcolor[gray]{.93}
TFM-RDI &15.8	&2.4	&27.8	&4.7	&10.1	&13.4	&12.4\\

\rowcolor[gray]{.93}
TFM-RDI+NCE    & 14.9 & 3.1 & 29.1 & 5.4 & 11.0 & 15.3 & 13.1 \\

\hline
\end{tabular}
\end{table*}

\section{Experiments}
\label{sec:experiments}
\subsection{Experimental Settings}
\subsubsection{Datasets}
Following previous studies~\cite{zhao2021success, byun2023introducing}, we conduct the experiments to evaluate the performance of our method on the widely used ImageNet-Compatible Dataset\footnote{\url{https://github.com/cleverhans-lab/cleverhans/tree/master/cleverhans_v3.1.0/examples/nips17_adversarial_competition/dataset}}. This dataset contains 1,000 images with the size $299 \times 299 \times 3$, and each image is provided with the real and target classes for targeted attacks. Additionally, we also utilize the CIFAR-10 dataset \cite{krizhevsky2009learning} to evaluate the targeted attacks against defensive models, following~\cite{byun2023introducing}.

\subsubsection{Source and target models} 
In our experiments, we use the ResNet-50, DenseNet-121, Inception-v3, and the adversarially trained ResNet-50 (adv-ResNet-50) \footnote{The adv-ResNet-50 was trained with adversarial samples constrained by small $l_2$-norm ($||\delta||_2 \leq 0.1$).} as the source models for crafting the targeted adversarial samples. To evaluate the transferability, we further employ ten pre-trained CNNs and five Transformer-based models as the target models. The ten CNNs are VGG-16, ResNet-18 (RN-18), ResNet-50 (RN-50), DenseNet-121 (DN-121), Xception (Xcep), MobileNet-v2 (MB-v2), EfficientNet-B0 (EF-B0), Inception ResNet-v2 (IR-v2), Inception-v3 (Inc-v3), and Inception-v4 (Inc-v4). The five Transformer-based models are ViT \cite{dosovitskiy2020image}, LeViT \cite{graham2021levit}, ConViT \cite{d2021convit}, Twins \cite{chu2021twins}, and PiT \cite{heo2021rethinking}. 

Moreover, on the CIFAR-10 dataset, we further employ four ensemble-based defensive models comprising three ResNet-20 networks (ens3-RN-20). These models are trained by using defensive settings: standard training, ADP \cite{pang2019improving}, GAL \cite{kariyappa2019improving}, and DEVRGE \cite{yang2020dverge}.

\subsubsection{Baselines} Following \cite{byun2023introducing}, we compose the different baseline attacks, encompassing DI-FGSM (DI)~\cite{xie2019improving}, RDI-FGSM (RDI)~\cite{zou2020improving}, TI-FGSM (TI) \cite{dong2019evading}, MI-FGSM (MI)~\cite{dong2018boosting}, VT-FGSM (VT)~\cite{wang2021enhancing}, SI-NI-FGSM (SI)~\cite{lin2020nesterov}, Admix~\cite{wang2021admix}, and ODI~\cite{byun2022improving}. We apply the MI and TI techniques across all attack methods, and for brevity, we omit `MI-TI' when denoting them. 

Following the baseline in the CFM~\cite{byun2023introducing}, we adopt RDI as our main baseline.  Specifically, we set the scale multipliers of image sizes to $1\sim\frac{330}{299}$ and $\frac{340}{299}$ for DI and RDI, respectively. For DI, we set the transformation probability to $p=0.7$. For TI, we set the convolutional kernel size to $k = 5$. For MI, we set the decay factor to $\mu=1.0$. For VT and SI, we set the number of samples and scales to 5, and $\beta$ of VT to 1.5. For Admix, we set the number of copies to $m_1 = 1$, the sample
number to $m_2 = 3$, and the admix ratio $\eta = 0.2$. 
For both Admix and CFM, we maintain a consistent batch size of 20, ensuring a fair comparison.

\subsubsection{Implementation Details} 
The training settings are mainly following~\cite{zhao2021success,byun2023introducing}. The adversarial perturbation is bounded by the commonly used $L_\infty \leq \frac{16}{255}$, and the step size for the iterative optimization is $\eta=\frac{2}{255}$. The total number of training iterations is set to 300. For our truncated feature mixing (TFM), we follow the settings in ~\cite{byun2023introducing}, that set and the feature mixing probability for each layer to 0.1 and the channel-wise
mixing ratio $\alpha$ to be randomly sampled from $U(0, 0.75)$. The temperature factor in Eq.~\ref{eq:ncl} is set to 2 for all source models. The targeted attack success rates after the final 300 iterations are reported for comparison. 


\subsection{Attacks on the ImageNet-Compatible Dataset}
In this part, we evaluate the performance of targeted attacks on the ImageNet-Compatible Dataset. We employ four pre-trained models from the ImageNet dataset as the source models to generate adversarial examples, and reported the corresponding performance in Table~\ref{tab:non-robust} and Table~\ref{tab:vit}.  

\subsubsection{Performance on Normal-Trained CNN Models}

The resulting targeted attack success rates against ten non-robust models are reported in Table~\ref{tab:non-robust}. Based on the results, we can have the following conclusions:
\textbf{1)} TFM-RDI consistently outperforms all baseline methods across all source models when using the same Logit loss for training. Notably, when compared to CFM-RDI, which employs stored clean image features for mixing, the proposed technique of truncated feature mixing demonstrates superior performance. For instance, when adv-RN-50 is utilized as the source model, the average targeted attack success rates increase from 78.4\% to 80.4\%.
\textbf{2)} The incorporation of the NCE loss component yields further improvements in the targeted transferability of adversarial examples for the four source models. In comparison to TFM-RDI, the average targeted attack success rates of TFM-RDI+NCE exhibit enhancements of 5.2\%, 4.6\%, and 7.5\% for RN-50, adv-RN-50, and DN-121, respectively. On the other hand, we notice that TFM-RDI+NCE has a similar performance to TFM-RDI for Inc-v3.

\subsubsection{Performance on Adversarially-Trained and Transformer-Based Models}
In this part, we extend our evaluation to encompass an adversarially trained model and five transformer-based models to further substantiate the effectiveness of our proposed methods. As depicted in Table~\ref{tab:vit}, it is evident that the novel truncated feature mixing technique consistently outperforms all baseline methods, including the present state-of-the-art CFM-RDI approach. Moreover, our introduced NCE loss function yields additional enhancements in targeted transferability against robust model and transformer-based models.

Specifically, when compared to CFM-RDI, the TFM-RDI and NCE techniques demonstrate a substantial improvement in the average targeted attack success rate, elevating it from 8.9\% to 12.4\% and 13.1\% when using the Inc-v3 as the source model. These findings strongly suggest that mixing truncated features and the proposed NCE loss both contribute significantly to the advancement of targeted transferability.

\subsubsection{Targeted attack success rates based on the number of iterations}

\begin{figure}[t]
    \centering
    \includegraphics[width=\linewidth]{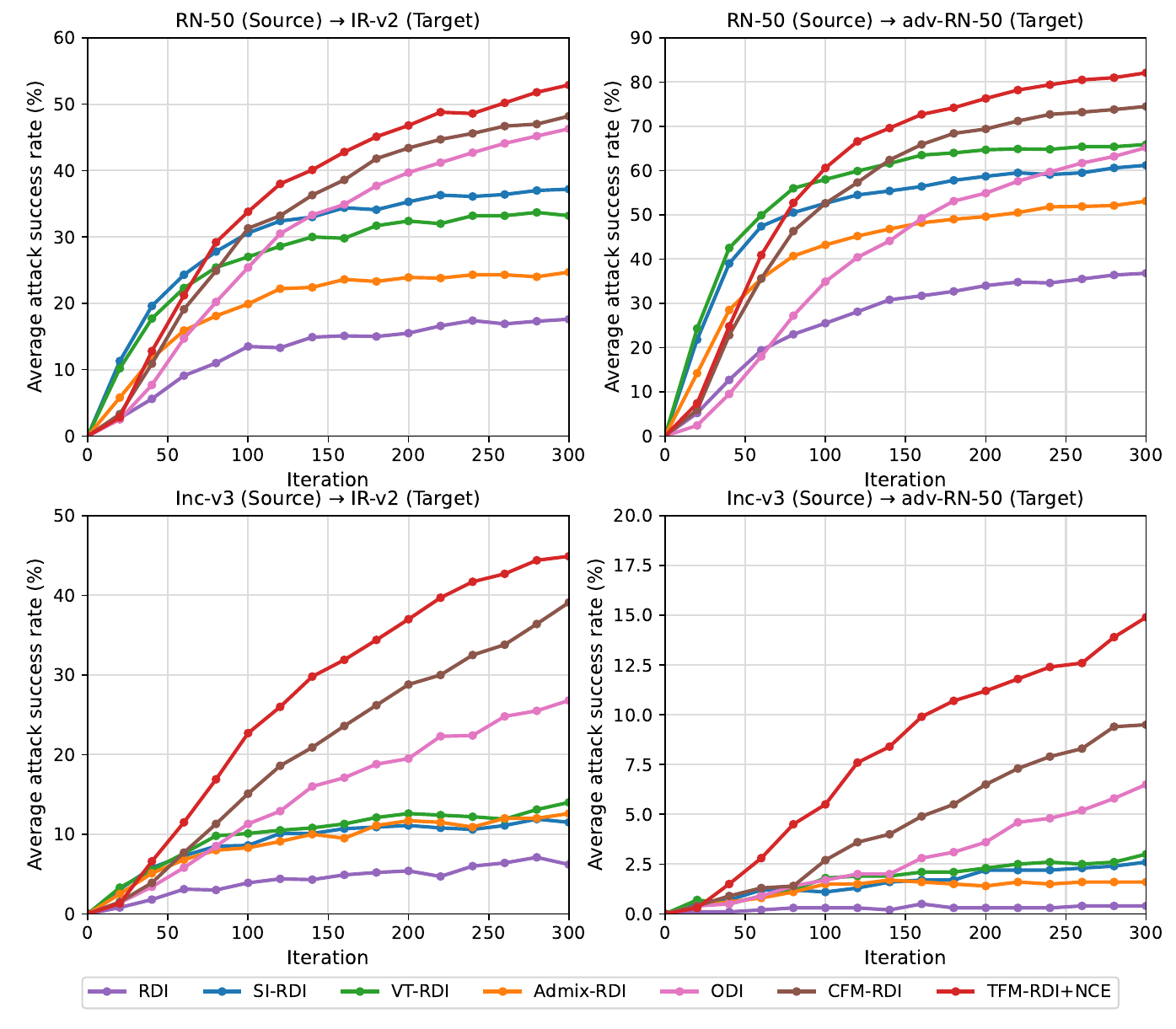}
    \caption{Targeted attack success rates (\%) based on the number
of iterations. Best viewed in color.}
    \label{fig:iter}
\end{figure}

In Figure~\ref{fig:iter}, we plot the average attack success rate according to training iterations of two different source models, \textit{i.e.}, ``ResNet-50 (RN-50)  and ``Inception-v3 (Inc-v3). The ``adv-ResNet-50 (adv-RN-50)" and  ``Inception ResNet-v2 (IR-v2)"  are used as the targeted models. From the figure, we can observe that our TFM with NCE loss outperforms other methods in average attack success rates when optimized for around 50 iterations. Besides, it requires more iterations to reach the saturation status, which further can verify our proposed loss function can alleviate the vanishing gradient issue in targeted attack as discussed in \cite{zhao2021success} and \cite{weng2023logit}.

\begin{table*}[ht!]
\centering
\caption{Targeted attack success rates (\%) and average computation time against nine target models, including four ensemble-based defensive models on the CIFAR-10
dataset.}
\label{tab:cifar10}
\begin{tabular}{lccccccccccc}
\bottomrule[1.2pt]
\textbf{Source: RN-50} & \multicolumn{9}{c}{Target model} &  \\
\cline{2-10}
\multirow{2}{*}{Attack} & \multirow{2}{*}{VGG-16} & \multirow{2}{*}{RN-18} & \multirow{2}{*}{MB-v2} & \multirow{2}{*}{Inc-v3} & \multirow{2}{*}{DN-121} & \multicolumn{4}{c}{ens3-RN-20} & \multirow{2}{*}{Avg.} &   \multirow{2}{*}{\makecell{Computation time\\ per image (sec)}}\\
 &  &  &  &  &  & Baseline & ADP & GAL & DVERGE \\
\hline
DI &66.4 &71.5 &62.7 &71.1& 84.2& 77.9& 56.5
&14.3 &15.6& 57.8  & 0.48\\
RDI& 66.4& 70.9& 64.1 &73.4 &82.8 &76.3 &55.8 &13.5 &14.9 &57.6 & 0.50\\
SI-RDI &72.9 &76.3 &77.1& 77.0& 84.7& 81.2 &65.5 &20.0 &22.4 &64.1 & 2.46\\
VT-RDI& 89.8& 87.1& 92.6& 92.9 &93.7& 94.4 &82.3 &24.3 &31.3& 76.5 & 2.96\\
Admix-RDI &74.2 &78.8 &76.2& 82.7& 89.2& 85.2 &66.4 &17.3 &18.4 &65.4 & 1.50\\
CFM-RDI &98.3& 97.7& 99.0& 99.0& 99.2& 98.8& 97.2 &54.9& 59.3 &89.3 & 0.55\\
\rowcolor[gray]{.93}
TFM-RDI &98.3&	97.9	&99.2&	98.7&	99.2	&98.3&	97.4&	61.2	&69.9&	91.1 & 0.58\\
\rowcolor[gray]{.93}
TFM-RDI+NCE &98.2&	97.5	&98.4	&98.8&	99.2	&98.5	&97.9&	62.0	&71.0	&91.3 & 0.59\\
\bottomrule[1.2pt]
\end{tabular}
\end{table*}

\subsection{Attack the CIFAR-10 Dataset}
To further validate the efficacy of our proposed method on a small-sized dataset, we extend our analysis to encompass transfer-based targeted attacks using the CIFAR-10 dataset. The targeted attack success rates against five non-robust models and four ensemble-based defensive models that consist of three ResNet-20 networks (ens3-RN-20) are presented in Table~\ref{tab:cifar10}. Notably, due to the saturated attack success rates on the five non-robust models and two robust models trained using the Baseline and ADP methods, the performances of CFM-RDI, TFM-RDI, and TFM-RDI+NCE are quite similar. However, for robust models trained with GAL and DEVERGE, which pose a greater challenge for deception, the step-by-step integration of our proposed components leads to a progressive improvement in the targeted transferability of adversarial examples. For example, the targeted attack success rates of attacking DVERGE of our TFM-RDI and TFM-RID+NCE are 69.9\% and 71.0, which is significantly higher than the 59.3\% of CFM-RDI.  

In addition to the transfer success rates, we also provide insights into the computation time for crafting adversarial examples using various attack techniques in Table~\ref{tab:cifar10}. Notably, each iterative attack was executed using a single NVIDIA 2080Ti GPU. We can find that our method only adds a slightly additional cost when compared with CFM-RDI. 

\subsection{Evaluation on Normalized Logit Calibration}

In this section, we first compare the normalized logit calibration with other different logit methods and then evaluate the effectiveness of incorporating the proposed $\mathcal{L}_{NCE}$ into various existing baselines. Next, we conduct experiments to analyze the influence of using different $T$ in $\mathcal{L}_{NCE}$. 


\subsubsection{Comparison with existing logit calibration}
We conduct a comparison between normalized logit calibration and existing logit calibrations~\cite{weng2023logit} by using RDI as the baseline. In Table~\ref{tab:calibration}, we provide the average success rates of targeted attacks against all sixteen target models by using four different source models.


From the table, we can have the following observations: \textbf{1)} The Gaussian-based logit calibration method (\textit{i.e.}, $\hat{z}_i = \frac{z_i - \mu}{\sigma \times T}$) exhibits enhanced targeted attack success rates and can outperform the Logit loss approach. However, it has a low performance in comparison to Temperature-based and Margin-based calibrations. This is mainly due to that the deviation $\sigma$ will oversmooth the Logits when optimized for several iterations. \textbf{2)} For the existing logit calibration method~\cite{weng2023logit}, the Temperature-based calibration ($T=5$) is more suitable for RN-50 and adv-RN-50 as source models, $T=10$ works slightly well for DN-121, and the Margin-based calibration is chosen for Inc-v3. This highlights that different network architectures require varying logit calibration strategies, potentially undermining practical applicability when introducing a new model for training. 
\textbf{3)} Our proposed normalized logit calibration demonstrates greater compatibility across diverse source models, achieving superior performance that can outperform other different logit calibration methods.

\begin{table}[t]
\centering
\caption{Comparison with existing logit calibration methods. RDI is utilized as the baseline. The average targeted attack success rate (\%) against sixteen target models on the ImageNet-Compatible dataset.}
\label{tab:calibration}
\begin{tabular}{lcccc}
\hline
    & RN-50 & adv-RN-50 & DN-121 & Inc-v3\\ \hline
Logit  & 34.8	& 55.7	& 24.2	& 9.2   \\ 
T=5    & 37.3	& 58.8	& 25.5	& 9.3   \\ 
T=10   & 37.1	& 57.6	& 25.7	& 9.3   \\ 
Margin & 37.1	& 58.5	& 25.4	& 9.5   \\
Gaussian & 36.3 & 57.0  & 25.3  & 9.4   \\
\rowcolor[gray]{.93}
NCE    & 37.8	& 60.6	& 26.1  & 9.4\\
\hline
\end{tabular}
\end{table}

\subsubsection{Incorporating various existing attack techniques}
Besides, we conduct a comprehensive comparison between NCE and Logit loss across diverse existing attack techniques, including DI, RDI, SI-RID, VT-RDI, Admix-RDI, ODI, and CFM-RDI. 
As observed from the results reported in Table~\ref{tab:loss}, the NCE technique demonstrates compatibility with most of the existing attack methods in generating transferable adversarial examples compared to the Logit loss. For instance, when considering CFM-RDI as the baseline, the average targeted attack success rates exhibit improvements of 4.1\%, 3.6\%, 4.3\%, and 0.1\% for source models RN-50, adv-RN-50, DN-121, and Inc-v3, respectively. On the other aspect, we also find that the NCE may not perform well for SI-RDI across all source models. We argue the main reason is that the down-scale operation in SI-RDI will influence the magnitude of the feature.

\begin{table*}[ht!]
\caption{Incorporating normalized logit calibration into various existing attack techniques. The average targeted attack success rate (\%) against sixteen target models on the ImageNet-Compatible dataset.}
\centering
\label{tab:loss}
\begin{tabular}{lcccccccc}
\toprule[1.2pt]
         & \multicolumn{2}{c}{RN-50} & \multicolumn{2}{c}{adv-RN-50} & \multicolumn{2}{c}{DN-121} & \multicolumn{2}{c}{Inc-v3} \\
          \cline{2-9}
Attack    & Logit   & NCE   & Logit     & NCE   & Logit     & NCE       & Logit     & NCE  \\ \hline
DI        & 24.7 & 25.7 \textcolor{blue}{(+1.0)} & 50.9 & 54.2 \textcolor{blue}{(+3.3)} & 17.2 & 18.3 \textcolor{blue}{(+1.1)} & 8.0 & 8.2 \textcolor{blue}{(+0.2)} \\
RDI       & 34.8 & 37.7 \textcolor{blue}{(+2.9)} & 55.7 & 59.6 \textcolor{blue}{(+3.9)} & 24.2 & 25.3 \textcolor{blue}{(+1.1)} & 9.2  & 9.4 \textcolor{blue}{(+0.2)}\\
SI-RDI    & 47.5 & 47.2 \textcolor{red}{(-0.3)} & 59.5 & 57.8 \textcolor{red}{(-1.7)} & 29.9 & 29.7 \textcolor{red}{(-0.2)} & 11.2 & 10.6 \textcolor{red}{(-0.6)}\\
VT-RDI    & 44.5 & 48.3 \textcolor{blue}{(+3.8)} & 52.2 & 56.0 \textcolor{blue}{(+3.8)} & 31.7 & 34.5 \textcolor{blue}{(+2.8)} & 12.5 & 12.6 \textcolor{blue}{(+0.1)}\\
Admix-RDI & 42.2 & 45.7 \textcolor{blue}{(+3.5)} & 58.9 & 63.9 \textcolor{blue}{(+5.0)} & 31.1 & 33.8 \textcolor{blue}{(+2.7)} & 12.0 & 12.6 \textcolor{blue}{(+0.6)} \\
ODI       & 53.7 & 56.6 \textcolor{blue}{(+2.9)} & 63.1 & 67.4 \textcolor{blue}{(+4.3)} & 42.5 & 45.8 \textcolor{blue}{(+3.3)} & 20.2 & 20.7 \textcolor{blue}{(+0.5)} \\
CFM-RDI   & 58.2 & 62.4 \textcolor{blue}{(+4.2)} & 71.0 & 74.7 \textcolor{blue}{(+3.6)} & 49.2 & 53.5 \textcolor{blue}{(+4.3)} & 27.8 & 27.9 \textcolor{blue}{(+0.1)}\\
\bottomrule[1.2pt]
\end{tabular}
\end{table*}

\begin{table*}[ht!]
\caption{Targeted attack success rates(\%) against ten target models by using different $T$.}
\centering
\label{tab:non-robust-t}
\begin{tabular}{lccccccccccc}
\toprule[1.2pt]
\textbf{Source: RN-50} & \multicolumn{9}{c}{Target model} & \multicolumn{1}{l}{} \\
 \cline{2-11}
TFM-RDI+NCE & VGG-16 & RN-18 & RN-50 & DN-121 & Xcep & MB-v2 & EF-B0 & IR-v2 & Inc-v3 & Inc-v4 & Avg. \\ \hline
T=1 & 87.8 & 89.4 & 98.9 & 92.6 & 51.3 & 84.9 & 80.8 & 49.9 & 70.1 & 63.6 & 76.9 \\
T=2 & 91.0 & 93.8 & 100.0 & 95.9 & 56.1 & 89.9 & 85.8 & 52.9 & 72.5 & 68.3 & 80.6 \\
T=5 & 92.8 & 96.1 & 100.0 & 98.0 & 55.6 & 90.8 & 89.4 & 53.3 & 73.6 & 68.0 & 81.8 \\
\midrule[1.2pt]

\textbf{Source: adv-RN-50} & \multicolumn{9}{c}{Target model} & \multicolumn{1}{l}{} \\
 \cline{2-11}
TFM-RDI+NCE & VGG-16 & RN-18 & RN-50 & DN-121 & Xcep & MB-v2 & EF-B0 & IR-v2 & Inc-v3 & Inc-v4 & Avg. \\ \hline
T=1 & 80.3 & 90.1 & 93.4 & 91.9 & 69.3 & 86.9 & 85.8 & 68.3 & 79.2 & 71.5 & 81.7 \\
T=2 & 85.7 & 93.2 & 95.4 & 94.4 & 73.3 & 89.3 & 89.7 & 70.9 & 83.5 & 75.3 & 85.1 \\
T=5 & 87.4 & 96.4 & 97.9 & 96.6 & 75.6 & 93.3 & 93.0 & 74.5 & 87.8 & 77.5 & 88.0 \\
\midrule[1.2pt]

\textbf{Source: DN-121} & \multicolumn{9}{c}{Target model} & \multicolumn{1}{l}{} \\
 \cline{2-11}
TFM-RDI+NCE & VGG-16 & RN-18 & RN-50 & DN-121 & Xcep & MB-v2 & EF-B0 & IR-v2 & Inc-v3 & Inc-v4 & Avg. \\ \hline
T=1 & 80.7 & 83.6 & 87.2 & 99.0 & 43.6 & 68.2 & 72.2 & 46.5 & 63.2 & 57.6 & 70.2 \\
T=2 & 85.7 & 89.6 & 92.5 & 100.0 & 48.2 & 73.4 & 78.0 & 50.6 & 67.4 & 62.5 & 74.8 \\
T=5 & 84.2 & 88.6 & 93.0 & 100.0 & 46.8 & 70.3 & 76.5 & 47.4 & 66.6 & 59.8 & 73.3 \\
\midrule[1.2pt]
\textbf{Source:Inc-v3} & \multicolumn{9}{c}{Target model} & \multicolumn{1}{l}{} \\
 \cline{2-11}
TFM-RDI+NCE & VGG-16 & RN-18 & RN-50 & DN-121 & Xcep & MB-v2 & EF-B0 & IR-v2 & Inc-v3 & Inc-v4 & Avg. \\ \hline
T=1 & 21.2 & 26.5 & 26.8 & 40.6 & 36.5 & 25.0 & 39.9 & 46.0 & 96.0 & 48.4 & 40.7 \\
T=2 & 22.1 & 26.6 & 27.4 & 39.1 & 39.4 & 23.8 & 39.1 & 44.9 & 96.8 & 49.4 & 40.9 \\
T=5 & 22.3 & 26.7 & 29.9 & 40.0 & 38.1 & 24.1 & 39.3 & 46.5 & 98.5 & 49.9 & 41.5 \\
\bottomrule[1.2pt]
\end{tabular}
\end{table*}

\begin{table*}[ht!]
\caption{Targeted attack success rates(\%) against a robust model and five Transformer-based models by using different $T$.}
\centering
\label{tab:vit-t}
\begin{tabular}{lccccccc}
\toprule[1.2pt]
\textbf{Source: RN-50} & \multicolumn{6}{c}{Target model} & \multicolumn{1}{l}{} \\
 \cline{2-7}
TFM-RDI+NCE & \makecell{adv\\-RN-50} & ViT & LeViT & ConViT & Twins & PiT & Avg. \\ \hline
T=1 & 78.0 & 5.0 & 48.2 & 9.1 & 26.4 & 27.7 & 32.4 \\
T=2 & 82.1 & 4.5 & 52.4 & 11.1 & 26.9 & 30.6 & 34.6 \\
T=5 & 83.1 & 4.5 & 50.6 & 9.9 & 29.0 & 29.6 & 34.4 \\

\midrule[1.2pt]
\textbf{Source: adv-RN-50} & \multicolumn{6}{c}{Target model} & \multicolumn{1}{l}{} \\
 \cline{2-7}
TFM-RDI+NCE & \makecell{adv\\-RN-50} & ViT & LeViT & ConViT & Twins & PiT & Avg. \\ \hline
T=1 & 98.8 & 29.2 & 72.2 & 45.1 & 56.4 & 65.1 & 61.1 \\
T=2 & 99.6 & 32.6 & 76.7 & 47.5 & 59.4 & 68.1 & 64.0 \\
T=5 & 100.0 & 31.5 & 79.3 & 48.4 & 60.3 & 70.0 & 64.9 \\

\midrule[1.2pt]
\textbf{Source: DN-121} & \multicolumn{6}{c}{Target model} & \multicolumn{1}{l}{} \\
 \cline{2-7}
TFM-RDI+NCE & \makecell{adv\\-RN-50} & ViT & LeViT & ConViT & Twins & PiT & Avg. \\ \hline
T=1 & 47.5 & 3.4 & 38.1 & 7.9 & 20.2 & 23.0 & 23.4 \\
T=2 & 50.7 & 3.1 & 41.8 & 7.2 & 22.3 & 23.7 & 24.8 \\
T=5 & 49.6 & 3.2 & 38.7 & 7.2 & 20.9 & 23.2 & 23.8 \\
\midrule[1.2pt]
\textbf{Source: Inc-v3} & \multicolumn{6}{c}{Target model} & \multicolumn{1}{l}{} \\
 \cline{2-7}
TFM-RDI+NCE & \makecell{adv\\-RN-50} & ViT & LeViT & ConViT & Twins & PiT & Avg. \\ \hline
T=1 & 15.8 & 3.0 & 27.5 & 5.0 & 12.0 & 14.9 & 13.0 \\
T=2 & 14.9 & 3.1 & 29.1 & 5.4 & 11.0 & 15.3 & 13.1 \\
T=5 & 15.9 & 3.1 & 29.9 & 5.1 & 13.5 & 15.6 & 13.8 \\

\bottomrule[1.2pt]
\end{tabular}
\end{table*}

\begin{table*}[ht!]
\caption{Targeted attack success rates(\%) against ten target models after removing or adding different features.}
\centering
\label{tab:non-robust-rank}
\begin{tabular}{lccccccccccc}
\toprule[1.2pt]
\textbf{Source: RN-50} & \multicolumn{9}{c}{Target model} & \multicolumn{1}{l}{} \\
 \cline{2-11}
 & VGG-16 & RN-18 & RN-50 & DN-121 & Xcep & MB-v2 & EF-B0 & IR-v2 & Inc-v3 & Inc-v4 & Avg. \\ \hline
- Rank-1 & 91.0 & 93.8 & 100.0 & 95.9 & 56.1 & 89.9 & 85.8 & 52.9 & 72.5 & 68.3 & 80.6 \\
- 0.5$\times$Rank-1 & 90.7 & 93.4 & 100.0 & 94.7 & 54.8 & 88.0 & 84.9 & 51.7 & 71.1 & 66.2 & 79.6 \\
- Rank-1 \& Rank-2 & 91.2 & 93.1 & 100.0 & 95.7 & 55.1 & 89.0 & 85.6 & 53.0 & 70.7 & 66.6 & 80.0 \\
+ 0.5$\times$Rank-1 & 89.0 & 91.2 & 100.0 & 93.5 & 55.2 & 86.9 & 83.8 & 52.4 & 70.8 & 64.4 & 78.7 \\
+ Rank-1 & 87.8 & 89.8 & 99.9 & 92.9 & 55.5 & 85.8 & 82.9 & 50.8 & 68.8 & 63.3 & 77.7 \\
\midrule[1.2pt]

\textbf{Source: adv-RN-50} & \multicolumn{9}{c}{Target model} & \multicolumn{1}{l}{} \\
 \cline{2-11}
 & VGG-16 & RN-18 & RN-50 & DN-121 & Xcep & MB-v2 & EF-B0 & IR-v2 & Inc-v3 & Inc-v4 & Avg. \\ \hline
- Rank-1 & 85.7 & 93.2 & 95.4 & 94.4 & 73.3 & 89.3 & 89.7 & 70.9 & 83.5 & 75.3 & 85.1 \\
- 0.5$\times$Rank-1 & 84.2 & 92.3 & 95.6 & 93.3 & 72.7 & 88.5 & 88.0 & 69.6 & 83.0 & 72.8 & 84.0 \\
- Rank-1 \& Rank-2 & 84.6 & 93.0 & 96.5 & 94.3 & 72.9 & 90.4 & 88.7 & 71.9 & 83.2 & 75.3 & 85.1 \\
+ 0.5$\times$Rank-1 & 80.0 & 89.6 & 93.3 & 91.2 & 70.8 & 86.4 & 87.4 & 68.8 & 80.8 & 71.1 & 81.9 \\
+ Rank-1 & 79.0 & 88.3 & 92.8 & 90.0 & 68.4 & 85.6 & 85.8 & 66.6 & 78.2 & 70.0 & 80.5 \\
\midrule[1.2pt]

\textbf{Source: DN-121} & \multicolumn{9}{c}{Target model} & \multicolumn{1}{l}{} \\
 \cline{2-11}
 & VGG-16 & RN-18 & RN-50 & DN-121 & Xcep & MB-v2 & EF-B0 & IR-v2 & Inc-v3 & Inc-v4 & Avg. \\ \hline
- Rank-1 & 85.7 & 89.6 & 92.5 & 100.0 & 48.2 & 73.4 & 78.0 & 50.6 & 67.4 & 62.5 & 74.8 \\
- 0.5$\times$Rank-1 & 84.3 & 87.4 & 90.6 & 100.0 & 46.4 & 69.3 & 74.2 & 47.5 & 64.3 & 58.2 & 72.2 \\
- Rank-1 \& Rank-2 & 84.4 & 89.7 & 92.1 & 100.0 & 46.8 & 72.7 & 76.0 & 49.2 & 67.1 & 61.9 & 74.0 \\
+ 0.5$\times$Rank-1 & 83.1 & 87.1 & 89.9 & 100.0 & 47.6 & 73.2 & 74.8 & 48.7 & 66.0 & 59.5 & 73.0 \\
+ Rank-1 & 84.3 & 87.4 & 89.4 & 100.0 & 50.8 & 73.3 & 76.0 & 50.9 & 65.6 & 60.8 & 73.8 \\
\midrule[1.2pt]
\textbf{Source:Inc-v3} & \multicolumn{9}{c}{Target model} & \multicolumn{1}{l}{} \\
 \cline{2-11}
 & VGG-16 & RN-18 & RN-50 & DN-121 & Xcep & MB-v2 & EF-B0 & IR-v2 & Inc-v3 & Inc-v4 & Avg. \\ \hline
- Rank-1 & 22.1 & 26.6 & 27.4 & 39.1 & 39.4 & 23.8 & 39.1 & 44.9 & 96.8 & 49.4 & 40.9 \\
- 0.5$\times$Rank-1 & 18.9 & 22.6 & 25.2 & 34.4 & 30.1 & 18.1 & 35.7 & 40.5 & 97.3 & 43.5 & 36.6 \\
- Rank-1 \& Rank-2 & 22.7 & 26.9 & 28.0 & 40.7 & 38.7 & 24.3 & 40.8 & 45.8 & 97.6 & 52.3 & 41.8 \\
+ 0.5$\times$Rank-1 & 18.4 & 21.8 & 23.6 & 35.6 & 30.1 & 19.9 & 34.6 & 37.5 & 94.9 & 39.9 & 35.6 \\
+ Rank-1 & 19.3 & 22.6 & 25.3 & 33.6 & 31.0 & 21.7 & 34.2 & 37.0 & 92.1 & 39.3 & 35.6 \\
\bottomrule[1.2pt]
\end{tabular}
\end{table*}

\begin{table*}[ht!]
\caption{Targeted attack success rates(\%) against a robust model and five Transformer-based models after removing or adding different features.}
\centering
\label{tab:vit-rank}
\begin{tabular}{lccccccc}
\toprule[1.2pt]
\textbf{Source: RN-50} & \multicolumn{6}{c}{Target model} & \multicolumn{1}{l}{} \\
 \cline{2-7}
 & \makecell{adv\\-RN-50} & ViT & LeViT & ConViT & Twins & PiT & Avg. \\ \hline
- Rank-1 & 82.1 & 4.5 & 52.4 & 11.1 & 26.9 & 30.6 & 34.6 \\
- 0.5$\times$Rank-1 & 80.2 & 5.0 & 50.9 & 9.4 & 27.0 & 28.1 & 33.4 \\
- Rank-1 \& Rank-2 & 81.7 & 5.4 & 50.9 & 9.2 & 28.5 & 29.5 & 34.2 \\
+ 0.5$\times$Rank-1 & 79.8 & 6.0 & 49.5 & 10.8 & 26.9 & 28.7 & 33.6 \\
+ Rank-1 & 77.5 & 5.2 & 49.2 & 10.2 & 29.2 & 29.1 & 33.4 \\

\midrule[1.2pt]
\textbf{Source: adv-RN-50} & \multicolumn{6}{c}{Target model} & \multicolumn{1}{l}{} \\
 \cline{2-7}
 & \makecell{adv\\-RN-50} & ViT & LeViT & ConViT & Twins & PiT & Avg. \\ \hline
- Rank-1 & 99.6 & 32.6 & 76.7 & 47.5 & 59.4 & 68.1 & 64.0 \\
- 0.5$\times$Rank-1 & 99.9 & 30.7 & 74.8 & 46.4 & 57.0 & 65.3 & 62.3\\
- Rank-1 \& Rank-2 & 99.8 & 32.3 & 76.3 & 47.5 & 60.1 & 68.0 & 64.0 \\
+ 0.5$\times$Rank-1 & 99.5 & 28.7 & 72.9 & 44.5 & 54.1 & 62.9 & 60.4\\
+ Rank-1 & 99.3 & 28.3 & 72.3 & 44.5 & 53.7 & 62.2 & 60.0\\

\midrule[1.2pt]
\textbf{Source: DN-121} & \multicolumn{6}{c}{Target model} & \multicolumn{1}{l}{} \\
 \cline{2-7}
 & \makecell{adv\\-RN-50} & ViT & LeViT & ConViT & Twins & PiT & Avg. \\ \hline
- Rank-1 & 50.7 & 3.1 & 41.8 & 7.2 & 22.3 & 23.7 & 24.8\\
- 0.5$\times$Rank-1 & 47.4 & 3.3 & 38.2 & 6.2 & 20.5 & 22.5 & 23.0\\
- Rank-1 \& Rank-2 & 49.7 & 3.4 & 40.2 & 7.1 & 19.7 & 24.3 & 24.1\\
+ 0.5$\times$Rank-1 & 49.3 & 3.5 & 40.0 & 7.5 & 21.8 & 25.9 & 24.7\\
+ Rank-1 & 51.1 & 4.0 & 41.5 & 7.2 & 23.7 & 26.4 & 25.7\\

\midrule[1.2pt]
\textbf{Source: Inc-v3} & \multicolumn{6}{c}{Target model} & \multicolumn{1}{l}{} \\
 \cline{2-7}
 & \makecell{adv\\-RN-50} & ViT & LeViT & ConViT & Twins & PiT & Avg. \\ \hline
- Rank-1 & 14.9 & 3.1 & 29.1 & 5.4 & 11.0 & 15.3 & 13.1\\
- 0.5$\times$Rank-1 & 14.7 & 2.9 & 25.7 & 4.5 & 12.0 & 13.7 & 12.2\\
- Rank-1 \& Rank-2 & 15.0 & 2.4 & 26.7 & 4.0 & 11.6 & 15.0 & 12.4\\
+ 0.5$\times$Rank-1 & 15.7 & 2.6 & 25.6 & 3.8 & 11.3 & 12.6 & 11.9 \\
+ Rank-1 & 14.5 & 2.9 & 23.4 & 3.8 & 10.3 & 11.9 & 11.1 \\
\bottomrule[1.2pt]
\end{tabular}
\end{table*}

\subsubsection{The influence of using different $T$}

In this part, we compare the performance of using different $T$ in $C = \frac{\tilde{z}_1 - \mu}{\sigma \times T}$. The experimental results are shown in Table~\ref{tab:non-robust-t} and Table~\ref{tab:vit-t}. From the tables, we can have the following observations: \textbf{1)} When using $T=1$, the performance is significantly decreased when compared with $T=2$ and $T=5$. This means that $T$ plays a significant role in our normalized logit calibration. \textbf{2)} Comparing the results of $T=2$ and $T=5$, we find that $T=5$ has better performance when attacking ten CNN-based models, and $T=2$ performs better when attacking the adv-RN-50 and 5 Transformer-based models. In this study, we directly set $T=2$ for the comparison with other methods.

\subsection{Evaluation of Truncated Feature Mixing}

In this part, we further evaluate the performance of removing and adding different features, including:
\begin{align}
    \nonumber \bm{X}^{'} &= \bm{X} - 0.5 \times s_{1} \bm{u}_{1} \bm{v}_{1}^{T},  \\ 
    \nonumber \bm{X}^{'} &= \bm{X} - s_{1} \bm{u}_{1} \bm{v}_{1}^{T} - s_{2} \bm{u}_{2} \bm{v}_{2}^{T}, \\
    \nonumber \bm{X}^{'} &= \bm{X} + 0.5 \times s_{1} \bm{u}_{1} \bm{v}_{1}^{T}, \\
    \bm{X}^{'} &= \bm{X} + s_{1} \bm{u}_{1} \bm{v}_{1}^{T}.
\end{align}
The experimental results are reported in Table~\ref{tab:non-robust-rank} and Table~\ref{tab:vit-rank}. 


From the tables, we can find that: \textbf{1)} The overall best performance is achieved by only removing the Rank-1 feature. Besides, the performance of removing both `Rank-1\&Rank-2' is slightly better than removing `0.5$\times$Rank-1'. \textbf{2)} When considering the results of attacking the adv-RN-50 and five transformer-based models, the performance of removing Rank-1 and removing both `Rank-1\&Rank-2' are very similar.
\textbf{3)} When adding the `0.5$\times$ Rank-1' and `Rank-1', the targeted attack success rates are worse than removing `Rank-1' feature. These findings provide evidence that the Rank-1 feature has a stronger correlation with the training source models, and removing this feature can enhance the transferability.

\section{Conclusion}
\label{sec:conclude}
This paper primarily focuses on enhancing the transferability of targeted adversarial samples. First, we propose a normalized logit calibration method to calibrate the logits by jointly considering the logit margin and logits distribution. The corresponding normalized cross-entropy loss function can significantly improve transferability. Additionally, we introduce a truncated feature mixing method to further improve the performance, removing the Rank-1 feature of the clean sample for mixing. Extensive experiments conducted on the ImageNet-Compatible and CIFAR-10 datasets with various models and training baselines demonstrate the effectiveness of the two proposed components for improving the targeted transferability, surpassing state-of-the-art comparison baselines by a large margin.

\section*{Acknowledgement}
This work is supported by the National Natural Science Foundation of China (No.~62276221, No.~62376232), and the Natural Science Foundation of Fujian Province of China (No. 2022J01002).

\bibliographystyle{IEEEtran}
\bibliography{ref}

\begin{thebibliography}{10}
\providecommand{\url}[1]{#1}
\csname url@samestyle\endcsname
\providecommand{\newblock}{\relax}
\providecommand{\bibinfo}[2]{#2}
\providecommand{\BIBentrySTDinterwordspacing}{\spaceskip=0pt\relax}
\providecommand{\BIBentryALTinterwordstretchfactor}{4}
\providecommand{\BIBentryALTinterwordspacing}{\spaceskip=\fontdimen2\font plus
\BIBentryALTinterwordstretchfactor\fontdimen3\font minus \fontdimen4\font\relax}
\providecommand{\BIBforeignlanguage}[2]{{%
\expandafter\ifx\csname l@#1\endcsname\relax
\typeout{** WARNING: IEEEtran.bst: No hyphenation pattern has been}%
\typeout{** loaded for the language `#1'. Using the pattern for}%
\typeout{** the default language instead.}%
\else
\language=\csname l@#1\endcsname
\fi
#2}}
\providecommand{\BIBdecl}{\relax}
\BIBdecl

\bibitem{simonyan2014very}
K.~Simonyan and A.~Zisserman, ``Very deep convolutional networks for large-scale image recognition,'' in \emph{ICLR}, 2015.

\bibitem{he2016deep}
K.~He, X.~Zhang, S.~Ren, and J.~Sun, ``Deep residual learning for image recognition,'' in \emph{CVPR}, 2016.

\bibitem{ren2015faster}
S.~Ren, K.~He, R.~Girshick, and J.~Sun, ``Faster r-cnn: Towards real-time object detection with region proposal networks,'' \emph{NeurIPS}, 2015.

\bibitem{liu2016ssd}
W.~Liu, D.~Anguelov, D.~Erhan, C.~Szegedy, S.~Reed, C.-Y. Fu, and A.~C. Berg, ``{SSD}: Single shot multibox detector,'' in \emph{ECCV}, 2016, pp. 21--37.

\bibitem{long2015fully}
J.~Long, E.~Shelhamer, and T.~Darrell, ``Fully convolutional networks for semantic segmentation,'' in \emph{CVPR}, 2015.

\bibitem{ronneberger2015u}
O.~Ronneberger, P.~Fischer, and T.~Brox, ``U-net: Convolutional networks for biomedical image segmentation,'' in \emph{MICCAI}, 2015, pp. 234--241.

\bibitem{goodfellow2015explaining}
I.~J. Goodfellow, J.~Shlens, and C.~Szegedy, ``Explaining and harnessing adversarial examples,'' in \emph{ICLR}, 2015.

\bibitem{chen2019transferability}
X.~Chen, S.~Wang, M.~Long, and J.~Wang, ``Transferability vs. discriminability: Batch spectral penalization for adversarial domain adaptation,'' in \emph{ICML}, 2019, pp. 1081--1090.

\bibitem{dong2018boosting}
Y.~Dong, F.~Liao, T.~Pang, H.~Su, J.~Zhu, X.~Hu, and J.~Li, ``Boosting adversarial attacks with momentum,'' in \emph{CVPR}, 2018, pp. 9185--9193.

\bibitem{lin2020nesterov}
J.~Lin, C.~Song, K.~He, L.~Wang, and J.~E. Hopcroft, ``Nesterov accelerated gradient and scale invariance for adversarial attacks,'' in \emph{ICLR}, 2020.

\bibitem{xie2019improving}
C.~Xie, Z.~Zhang, Y.~Zhou, S.~Bai, J.~Wang, Z.~Ren, and A.~L. Yuille, ``Improving transferability of adversarial examples with input diversity,'' in \emph{CVPR}, 2019, pp. 2730--2739.

\bibitem{byun2022improving}
J.~Byun, S.~Cho, M.-J. Kwon, H.-S. Kim, and C.~Kim, ``Improving the transferability of targeted adversarial examples through object-based diverse input,'' in \emph{CVPR}, 2022, pp. 15\,244--15\,253.

\bibitem{li2020towards}
M.~Li, C.~Deng, T.~Li, J.~Yan, X.~Gao, and H.~Huang, ``Towards transferable targeted attack,'' in \emph{CVPR}, 2020, pp. 641--649.

\bibitem{zhao2021success}
Z.~Zhao, Z.~Liu, and M.~Larson, ``On success and simplicity: A second look at transferable targeted attacks,'' \emph{NeurIPS}, pp. 6115--6128, 2021.

\bibitem{weng2023logit}
J.~Weng, Z.~Luo, S.~Li, N.~Sebe, and Z.~Zhong, ``Logit margin matters: Improving transferable targeted adversarial attack by logit calibration,'' \emph{IEEE Transactions on Information Forensics and Security}, 2023.

\bibitem{byun2023introducing}
J.~Byun, M.-J. Kwon, S.~Cho, Y.~Kim, and C.~Kim, ``Introducing competition to boost the transferability of targeted adversarial examples through clean feature mixup,'' in \emph{CVPR}, 2023, pp. 24\,648--24\,657.

\bibitem{muhammad2020eigen}
M.~B. Muhammad and M.~Yeasin, ``Eigen-cam: Class activation map using principal components,'' in \emph{IJCNN}, 2020, pp. 1--7.

\bibitem{song2022rankfeat}
Y.~Song, N.~Sebe, and W.~Wang, ``Rankfeat: Rank-1 feature removal for out-of-distribution detection,'' \emph{NeurIPS}, pp. 17\,885--17\,898, 2022.

\bibitem{dong2019evading}
Y.~Dong, T.~Pang, H.~Su, and J.~Zhu, ``Evading defenses to transferable adversarial examples by translation-invariant attacks,'' in \emph{CVPR}, 2019, pp. 4312--4321.

\bibitem{weng2023exploring}
J.~Weng, Z.~Luo, Z.~Zhong, D.~Lin, and S.~Li, ``Exploring non-target knowledge for improving ensemble universal adversarial attacks,'' in \emph{AAAI}, 2023, pp. 2768--2775.

\bibitem{kurakin2018adversarial}
A.~Kurakin, I.~J. Goodfellow, and S.~Bengio, ``Adversarial examples in the physical world,'' in \emph{Artificial intelligence safety and security}, 2018, pp. 99--112.

\bibitem{wang2021enhancing}
X.~Wang and K.~He, ``Enhancing the transferability of adversarial attacks through variance tuning,'' in \emph{CVPR}, 2021, pp. 1924--1933.

\bibitem{zou2020improving}
J.~Zou, Z.~Pan, J.~Qiu, X.~Liu, T.~Rui, and W.~Li, ``Improving the transferability of adversarial examples with resized-diverse-inputs, diversity-ensemble and region fitting,'' in \emph{ECCV}, 2020, pp. 563--579.

\bibitem{wang2021admix}
X.~Wang, X.~He, J.~Wang, and K.~He, ``Admix: Enhancing the transferability of adversarial attacks,'' in \emph{ICCV}, 2021, pp. 16\,158--16\,167.

\bibitem{long2022frequency}
Y.~Long, Q.~Zhang, B.~Zeng, L.~Gao, X.~Liu, J.~Zhang, and J.~Song, ``Frequency domain model augmentation for adversarial attack,'' in \emph{ECCV}, 2022, pp. 549--566.

\bibitem{huang2019enhancing}
Q.~Huang, I.~Katsman, H.~He, Z.~Gu, S.~Belongie, and S.-N. Lim, ``Enhancing adversarial example transferability with an intermediate level attack,'' in \emph{CVPR}, 2019, pp. 4733--4742.

\bibitem{inkawhich2020transferable}
N.~Inkawhich, K.~J. Liang, L.~Carin, and Y.~Chen, ``Transferable perturbations of deep feature distributions,'' \emph{arXiv preprint arXiv:2004.12519}, 2020.

\bibitem{inkawhich2020perturbing}
N.~Inkawhich, K.~Liang, B.~Wang, M.~Inkawhich, L.~Carin, and Y.~Chen, ``Perturbing across the feature hierarchy to improve standard and strict blackbox attack transferability,'' \emph{NeruIPS}, vol.~33, pp. 20\,791--20\,801, 2020.

\bibitem{gao2021feature}
L.~Gao, Y.~Cheng, Q.~Zhang, X.~Xu, and J.~Song, ``Feature space targeted attacks by statistic alignment,'' \emph{arXiv preprint arXiv:2105.11645}, 2021.

\bibitem{krizhevsky2009learning}
A.~Krizhevsky, G.~Hinton \emph{et~al.}, ``Learning multiple layers of features from tiny images,'' \emph{Master's thesis, Department of Computer Science, University of Toronto}, 2009.

\bibitem{dosovitskiy2020image}
A.~Dosovitskiy, L.~Beyer, A.~Kolesnikov, D.~Weissenborn, X.~Zhai, T.~Unterthiner, M.~Dehghani, M.~Minderer, G.~Heigold, S.~Gelly \emph{et~al.}, ``An image is worth 16x16 words: Transformers for image recognition at scale,'' \emph{arXiv preprint arXiv:2010.11929}, 2020.

\bibitem{graham2021levit}
B.~Graham, A.~El-Nouby, H.~Touvron, P.~Stock, A.~Joulin, H.~J{\'e}gou, and M.~Douze, ``Levit: a vision transformer in convnet's clothing for faster inference,'' in \emph{ICCV}, 2021, pp. 12\,259--12\,269.

\bibitem{d2021convit}
S.~d’Ascoli, H.~Touvron, M.~L. Leavitt, A.~S. Morcos, G.~Biroli, and L.~Sagun, ``Convit: Improving vision transformers with soft convolutional inductive biases,'' in \emph{ICML}, 2021, pp. 2286--2296.

\bibitem{chu2021twins}
X.~Chu, Z.~Tian, Y.~Wang, B.~Zhang, H.~Ren, X.~Wei, H.~Xia, and C.~Shen, ``Twins: Revisiting the design of spatial attention in vision transformers,'' \emph{NeurIPS}, pp. 9355--9366, 2021.

\bibitem{heo2021rethinking}
B.~Heo, S.~Yun, D.~Han, S.~Chun, J.~Choe, and S.~J. Oh, ``Rethinking spatial dimensions of vision transformers,'' in \emph{CVPR}, 2021, pp. 11\,936--11\,945.

\bibitem{pang2019improving}
T.~Pang, K.~Xu, C.~Du, N.~Chen, and J.~Zhu, ``Improving adversarial robustness via promoting ensemble diversity,'' in \emph{ICML}, 2019, pp. 4970--4979.

\bibitem{kariyappa2019improving}
S.~Kariyappa and M.~K. Qureshi, ``Improving adversarial robustness of ensembles with diversity training,'' \emph{arXiv preprint arXiv:1901.09981}, 2019.

\bibitem{yang2020dverge}
H.~Yang, J.~Zhang, H.~Dong, N.~Inkawhich, A.~Gardner, A.~Touchet, W.~Wilkes, H.~Berry, and H.~Li, ``Dverge: diversifying vulnerabilities for enhanced robust generation of ensembles,'' \emph{NeurIPS}, pp. 5505--5515, 2020.

\end{thebibliography}
\end{document}